# Learning a Cross-modality Anomaly Detector for Remote Sensing Imagery

*Jingtao Li, Xinyu Wang\*, Hengwei Zhao, and Yanfei Zhong, Senior Member, IEEE*

***Abstract*—Remote sensing anomaly detector can find the objects deviating from the background as potential targets for Earth monitoring. Given the diversity in earth anomaly types, designing a transferring model with cross-modality detection ability should be cost-effective and flexible to new earth observation sources and anomaly types. However, the current anomaly detectors aim to learn the certain background distribution, the trained model cannot be transferred to unseen images. Inspired by the fact that the deviation metric for score ranking is consistent and independent from the image distribution, this study exploits the learning target conversion from the varying background distribution to the consistent deviation metric. We theoretically prove that the large-margin condition in labeled samples ensures the transferring ability of learned deviation metric. To satisfy this condition, two large margin losses for pixel-level and feature-level deviation ranking are proposed respectively. Since the real anomalies are difficult to acquire, anomaly simulation strategies are designed to compute the model loss. With the large-margin learning for deviation metric, the trained model achieves cross-modality detection ability in five modalities—hyperspectral, visible light, synthetic aperture radar (SAR), infrared and low-light—in zero-shot manner.

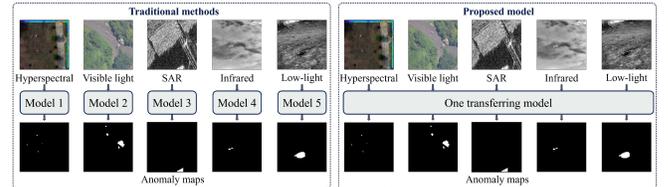

Fig. 1. The cross-modality detection paradigm of proposed model. Given the modalities with different imaging mechanisms and huge distribution difference, traditional models need to be trained for each modality while proposed model can infer the unseen modalities directly with zero-shot transferring ability.

## 1. Introduction

Remote sensing images can be used to monitor anomalies on the Earth's surface in a large-scale and consistent space [1]. Anomaly detection in remote sensing (ADRS) task aims to find the pixels deviating from the background spectrally or spatially, which are detected without any prior knowledge [2], [3], [4]. The anomalies vary in category and electromagnetic response. For example, landslide anomalies exhibit a response in the visible and radar range, while fire anomalies are mainly related to the thermal infrared spectra [5]. Given the diversity in anomaly types and responses across modalities, building a transferring model with cross-modality detection ability for ADRS (as Fig. 1) to different modalities would be cost-effective and allow easy adaptation to new data sources and anomaly types.

Designing a such transferring model is challenging due to the difference in imaging mechanisms of different modalities. Specifically, the hyperspectral modality can record a continuous spectrum from visible to short-wave infrared [6], and thus the acquired imagery always has hundreds of channels for precise recognition [7], [8]. In contrast, the synthetic aperture radar (SAR) modality is a side-looking radar that records the received echoes coherently [9], [10], providing more structural information with several channels. Besides, large-scale scenes encompass diverse backgrounds, including forests, urban areas, and oceans, with highly variable distributions [11], [12].

Since the huge distribution difference, most anomaly detectors are still limited to a single modality since they aim to learn the certain background distribution for each image. They focus on describing the background distribution with a statistical-based [13], [14], [15], [16], representation-based [17], [18], [19], [20], or deep learning based method [21], [22], [23], [24] first, and then use some deviation metric directly to obtain the anomaly score. The statistics-based methods describe the background distribution with some statistical model (e.g., multivariate Gaussian distribution) [13]. The representation-based methods describe the background with a hand-crafted dictionary considering the low-rank and sparsity priors [2], [25]. The deep learning based models mostly use reconstruction models to learn the background distribution and assume that the normal pixels have a smaller reconstruction error than the anomaly ones [26], [27], [28]. After obtaining the background distribution, some deviation metric such as the Mahalanobis distance [15] and the mean squared error [29] is used directly to obtain the anomaly score. However, the background distribution always varies in unseen images and thus the prior constructed detector for certain background is not applicable any more. This is the main reason why the existing models need to be constructed again for each image and lack the cross-modality transferring ability.

To solve the transferring problem, finding an image-

Jingtao Li, Hengwei Zhao, and Yanfei Zhong are with the State Key Laboratory of Information Engineering in Surveying, Mapping and Remote Sensing and the Hubei Provincial Engineering Research Center of Natural Resources Remote Sensing Monitoring, Wuhan University, Wuhan 430072, China. (e-mail: jingtaoli@whu.edu.cn; whu_zhaohw@whu.edu.cn; zhongyanfei@whu.edu.cn).

Xinyu Wang is with the School of Remote Sensing and Information Engineering, Wuhan University, Wuhan 430072, China (*Corresponding author, e-mail: wangxinyu@whu.edu.cn).

independent learning target is the core step. We observe that although the modality and scene have changed, most detectors use fixed deviation metrics (e.g., Mahalanobis distance) [15], [30] to compute the anomaly score. The learned background distribution act as the varying input for the deviation metric while the deviation metric itself is unchanged and image-independent. Inspired by this, we exploit to learn the deviation metric directly, which accepts the original image as input and ranks the deviation degree for each pixel. Different from the hand-crafted metrics, our score process is end-to-end without the need to obtain the background distribution first.

In this study, we build a cross-modality detector by learning an image-independent deviation metric. Instantiating the deviation metric as a learned deep model, we first theoretically prove that although the cross-modality images may be unseen at training stage, once the trained model can meet the large margin condition in the limited training samples, it can also rank the unseen anomaly and background correctly. Based on the proved Theorem, two large-margin deviation ranking losses are further proposed at pixel-level and feature-level. The pixel-level loss is derived from the common ranking metric (Area Under the Curve) AUC, and thus has a smaller gap between the optimization and the evaluation. The feature-level loss is designed to optimize the ranking of features with the hypersphere centers. Both the pixel-level and feature-level losses punish the small margin even for the correct ranking. Since the real anomalies are difficult to acquire, the anomaly simulation strategy is proposed to generate labeled anomalies and compute the large-margin losses.

In brief, the main contributions of this paper can be summarized as follows.

(1) The anomaly detection model with cross-modality transferring ability is built by converting the learning target from the certain background distribution to the image-independent deviation metric.

(2) We theoretically prove that meeting the large margin condition in training samples can guarantee the correct deviation rank for unseen anomaly and background.

(3) The large-margin ranking losses at pixel-level and feature-level are designed. The losses work together with simulated samples and punish the small margin even for the correct ranking.

The rest of this paper is organized as follows. Section 2 introduces the related work in remote sensing anomaly detection. Section 3 provides a detailed description of the motivation and the learning method of deviation metric. Section 4 gives the experimental results and analysis. Finally, the paper is concluded in Section 5. The code is available at https://github.com/Jingtao-Li-CVer/UniADRS.

## 2. RELATED WORK

### 2.1 Anomaly Detection Task in Remote Sensing

ADRS involves finding the objects that are anomalous to the background, without any prior information [31]. There is not an unambiguous way to define an anomaly, which is generally identified as an observation deviating from the background, spectrally or spatially [2], [4]. In fact, the category of the anomalies depends on the particular application. The anomalies can be the camouflage [44] or vehicles in military surveillance [32], rare minerals in geological detection [33], infected trees in forestry [30], and ships on the sea [33]. Since the ADRS methods do not use any prior knowledge, they cannot distinguish between real anomalies and detections that are not of interest. The detection result is often a first step, which provides the potential targets for the subsequent recognition [34].

Some fields may seem similar to the ADRS methods, but there are significant differences. Anomaly detection in medical or industrial images finds the anomaly pattern given a set of normal samples [35], where the normal pattern is no longer the background defined in the ADRS. The detected anomalies have both large and small areas. Despite some researchers having defined the normal pattern as the same as the industrial one in high-resolution optical images [30], we inherit te classical anomaly definition in the remote sensing community and treat the background as the normal pattern in each scene. Compared to tiny object detection [36], [37], the ADRS task is unsupervised without preset categories and labeled training samples. In addition, the anomalies in an ADRS are always small and rare, while tiny object detection also considers abundant small objects (e.g., cars in a parking lot).

### 2.2 Anomaly Detection Methods in Remote Sensing

Since the difficulty to acquire the real anomalies, most ADRS methods aim to extract the discriminative background features first and then use a distance metric to assign the anomaly score for each pixel. According to the principles of background learning, the detection models can be divided into three categories: statistical-based [13], [14], [15], [16], [38], [39] models, representation-based [17], [18], [19], [20] models, and deep learning based method [21], [22], [23], [24], [40], [41].

**Statistics-based models.** This statistical models aim to describe the background distribution with statistical techniques [14], where the likelihood implies the anomaly degree. For example, the classic Reed-Xiaoli (RX) detector models the background as a multivariate Gaussian distribution [15]. The Mahalanobis distance between the test pixel and the modeled distribution is then treated as the anomaly degree. Inspired by the RX detector, many improved variants have been proposed, such as the kernel RX-AD [42], weighted-RX-AD and linear filter-based RX-AD [43] and spectral-spatial feature extraction-based AD [44]. Recently, Chang proposed [45], [46] a target-to-anomaly conversion mechanism, which converts many well-known target detectors to the corresponding anomaly versions. Except for the accuracy improvement, some researchers have focused on real-time processing with RX detectors [38], [47], [48]. To address the difficulty of determining the distribution form, statistical cluster centers and decision hyperspheres have also been deployed [49], [50]. For the SAR modality, Haitman et al. [51] used both the RX detector and the non-negative matrix factorization (NNMF) learning algorithm to detect the sludge pools in Israel. Despite the statistical methods having a clear mathematical basis, the constructed distribution is only suitable for single images [27] and does not have the ability to be cross-modal or cross-scene.

**Representation-based models.** The representation-based models construct the detector considering the prior properties of the anomalies or background [17], [52], and include sparsity, collaborative, and low-rank based detectors. Ling *et al.* [18] constructed a sparsity-based detector with the sum-to-one and non-negativity constraints, making the detector less sensitive to the anomalies. Differing from sparse representation, collaborative representation assumes that the background pixels can be reconstructed by the surrounding pixels while the anomalies cannot [19]. The classic collaborative representation detector (CRD) follows this assumption [20]. To make full use of the global structural information (i.e., low-rank property), the low-rank and sparse matrix decomposition model (LSDM) was designed by decomposing the hyperspectral image into a low-rank background and sparse anomalies [17]. Sun *et al.* [53] implemented the LSDM technique with robust principal component analysis (RPCA) [54]. Zhang *et al.* [55] proposed a detector based on the low-rank and sparse matrix decomposition (LRaSMD) technique and applied the Mahalanobis distance to estimate the background part (LSMAD). Xu *et al.* [56] first introduced the background dictionary and proposed a detector based on low-rank and sparse representation (LRASR). Although the representation models do not rely on specific statistical distribution, the used background dictionary needs to be constructed for each modality and scene, limiting the transferring ability.

**Deep learning based models.** Most deep learning based models follow a two-step paradigm [27], [57], where they assume that the normal pixels have a smaller reconstruction error with the deep model than the anomaly ones. Li *et al.* [40] first introduced a convolutional neural network (CNN) into the hyperspectral anomaly detection (HAD) task in a supervised way. To detect anomalies according to a practical situation, some unsupervised methods have been proposed. For example, Xie *et al.* [23] proposed the spectral constrained adversarial autoencoder (SC_AAE), where a spectral constraint strategy is incorporated for better latent representation. However, these methods always involve complicated manual parameter setting and preprocessing steps. To this end, Wang *et al.* [29] proposed the autonomous hyperspectral anomaly detection network (Auto-AD) with an adaptive-weighted loss function, where a high reconstruction error implies anomaly. Except for the autoencoder model, generative adversarial network (GAN)-based models have also been used, where the generation error from real images is treated as the anomaly degree [58], [59], [60]. For example, Jiang *et al.* [60] introduced a semi-supervised GAN with dual RX detector to learn the discriminative reconstruction of background and anomalies. Inspired by the fact that both the autoencoder-based models and GAN-based models adopt the reconstruction proxy task and need to be trained for each image, Li *et al.* [27] proposed the one-step detection paradigm and transferred direct detection (TDD) model, where the proxy task is abandoned and the trained model can be transferred to unseen images directly. However, the TDD model is still limited in the hyperspectral modality due to the proxy classification optimization and the simulated spectral anomalies.

**Fast anomaly detection.** Since anomalies may appear in a short time and bring huge losses, many efforts have been made to improve the detection speed. Chen *et al.* [61] designed causal processing with Kalman filters and achieved real-time performance. To better conform to the push-broom scanners, Díaz *et al.* [62] proposed a line-by-line anomaly detector (LbL-FAD), which used hardware-friendly alternative to compute the orthogonal subspace spanned by selected background pixels to make the anomalies easily separated. López-Fandiño *et al.* [63] designed a parallel algorithm to be executed on multi-node heterogeneous computing platforms based on Reed–Xiaoli (RX) [15]. A *et al.* [64] proposed a fast local RX (FLRX) detector to achieve near-time performance. It can be seen that most fast detectors are based on statistical models to achieve real-time performance. Although prior work has used field programmable gate array (FPGA) to speed up the deep anomaly detectors for multispectral imagery [65], their transferability is still limited in known scenes due to the learning target of certain background. In hyperspectral community, the paradigm of training and testing on each image prevents the deep model from being real-time. This study tackles the problem by transforming the learning target from varying background to fixed deviation ranking, eliminating the training time for fast processing on unseen images.

## 3. A TRANSFERRING MODEL FOR REMOTE SENSING ANOMALY DETECTION

In this section, we first formulate the ADRS task and clarify the motivation in Section 3.1, where our main idea is to change the learning target from the varying image distribution to the image-independent deviation metric (Fig. 2). The analysis in Section 3.2 shows that satisfying the large margin condition in the labeled samples is the key for the transferring ability of learned deviation metric. To satisfy the condition, two large-margin losses are proposed in pixel-level and feature-level respectively for the correct deviation ranking in Section 3.3. Since real anomalies are difficult to acquire, we design the anomaly simulating strategies in Section 3.4 for computing the deviation ranking loss. Fig. 3 gives an overview of the built transferring model.

### 3.1 Motivation: From Single to Cross-Modality Detection

Given a remote sensing image $\mathbf{X} \in R^{H \times W \times C}$, $\mathbf{X} = \mathbf{B} + \mathbf{A}$ in the ideal condition without noise, where $\mathbf{B}$ is the background and $\mathbf{A}$ is the anomaly component. In ADRS task, $\mathbf{A}$ is always the small target with the empirical ratio in range [0.0019%, 0.48%] obtained by statistics of 12 well-known datasets from [66], [67], [68], [69], [70], [71], [72], [73], [74].

Regardless of the instantiation difference, current detection models rely on both the image distribution $P(\mathbf{X})$ and the deviation metric function $S$, which measures the scalar deviation degree of the pixel $x \in \mathbf{X}$ and the $P(\mathbf{X})$. The deviation degree for each pixel reflects its occurrence probability and the distribution difference with the whole image. Ideally, any anomaly pixel $\boldsymbol{a} \in \mathbf{A}$ and any background pixel $\boldsymbol{b} \in \mathbf{B}$ should satisfy the ranking inequality $S(\boldsymbol{a}, P(\mathbf{X})) \geq S(\boldsymbol{b}, P(\mathbf{X}))$, implying the higher deviation and anomaly score of $\mathbf{A}$ than $\mathbf{B}$. For example, RXD instantiates the $P(\mathbf{X})$ as the multivariate Gaussian distribution and instantiates the $S$ as the Mahalanobis distance [15]. LRASR [56] instantiates the $P(\mathbf{X})$ as a background dictionary and instantiates the $S$ as the reconstruction error.

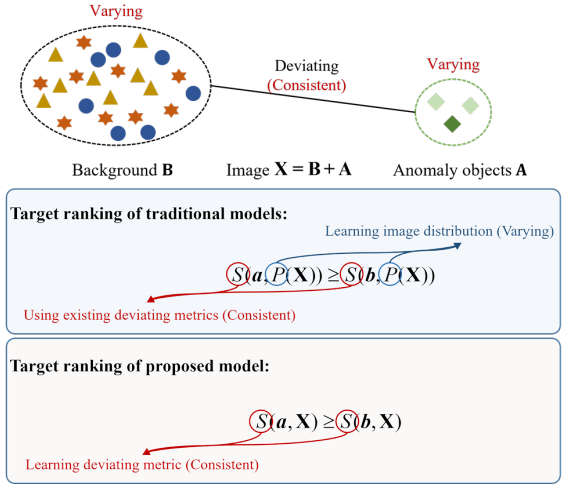

Fig. 2. Description of our main principle difference compared to traditional models. Traditional models focus on learning the certain image distribution first and then use some existing deviating metric to rank the anomaly score. In different modalities, since the distributions of background and anomaly are varying, the prior learned model cannot be transferred to unseen image distribution. Inspired by the fact that the deviating metric $S$ is independent and consistent for all the modality image, proposed model aims to bypass the image distribution learning and learn the deviating metric directly, achieving the cross-modality detection ability.

We observe that most models only concern about the quality of $P(\mathbf{X})$ and use some existing distance metric directly to get the final anomaly map (e.g., Mahalanobis distance). They mainly differ in the methods of representing $P(\mathbf{X})$ (statistical-based, representation-based or deep learning-based). However, when $P(\mathbf{X})$ has changed given an unseen image of different modality, the model needs to be rebuilt or trained, lacking the transferring detection ability.

To solve the transferring problem and increase the flexibility of the detection model, ***our main idea is to abandon the learning target of $P(\mathbf{X})$ but learn the deviation metric $S$ directly***. Since $S$ is independent of the $\mathbf{X}$, the learned $S$ can be consistent given any unseen image, thus achieving the cross-modality transferring detection.

*3.2 Learning Deviation Metric for Correct Ranking*

We made $S$ learnable by instantiating it as a trained deep model. The expected $S$ can score the deviation degree for each pixel and satisfy the ranking inequality $S(\boldsymbol{a}, \mathbf{X}) \geq S(\boldsymbol{b}, \mathbf{X})$. Different from the traditional deviation metric such as the Mahalanobis distance, our $S$ does not need to obtain the $P(\mathbf{X})$ first. Fig. 2 shows our main difference compared to the traditional models. For simplicity, unless otherwise specified, $S(\boldsymbol{a}, \mathbf{X})$ is shortened to $S(\boldsymbol{a})$ and $S(\boldsymbol{b}, \mathbf{X})$ is shortened to $S(\boldsymbol{b})$ in the following paragraphs.

To answer the question of "***how to ensure the learned deviation ranking ability of S transferring?***". In this section, we theoretically prove that the transferring ability of $S$ can be achieved once meeting the large margin condition in limited labeled samples (statement is provided in Theorem 1).

**Theorem 1.** Set $Q_l$ be the training set of many labeled samples (the anomaly pixel $\boldsymbol{a}_j \in R^C$ indexed by $j$, the background pixel $\boldsymbol{b}_i \in R^C$ indexed by $i$). Set $\delta$ be the smallest radius, such that for any unseen anomaly pixel $\boldsymbol{u}_a$ or the unseen background pixel $\boldsymbol{u}_b$, $\boldsymbol{u}_a$ is in the $\delta$-ball of some $\boldsymbol{a}_j$ in $Q_l$ and $\boldsymbol{u}_b$ is in the $\delta$-ball of some $\boldsymbol{b}_i$ in $Q_l$. If the score function $S$ meets the $\lambda_s$-Lipschitz continuous condition and has correctly ranked the $Q_l$ with a large margin, i.e., $S(\boldsymbol{a}_j) - S(\boldsymbol{b}_i) \geq 2\delta\lambda_s$ holds for all the labeled pixels, then $S$ can also rank the unseen pixels of different modality correctly, i.e., $S(\boldsymbol{u}_a) \geq S(\boldsymbol{u}_b)$.

**Proof.** Considering the $\lambda_s$-Lipschitz continuous property of $S$ and the $\boldsymbol{u}_a$ and $\boldsymbol{u}_b$ are assumed to be close to $\boldsymbol{a}_j$ and $\boldsymbol{b}_i$ respectively with the distance smaller than $\delta$. Eq. (1) and Eq. (2) can be obtained.

$$S(\boldsymbol{a}_j) - \delta\lambda_s \leq S(\boldsymbol{u}_a) \quad (1)$$
$$-S(\boldsymbol{b}_i) - \delta\lambda_s \leq -S(\boldsymbol{u}_b) \quad (2)$$

Adding the inequalities (1) and (2), and with the condition $S(\boldsymbol{a}_j) - S(\boldsymbol{b}_i) \geq 2\delta\lambda_s$, we can further obtain the Eq. (3). Thus, $S(\boldsymbol{u}_a) \geq S(\boldsymbol{u}_b)$ can hold.

$$0 \leq S(\boldsymbol{a}_j) - S(\boldsymbol{b}_i) - 2\delta\lambda_s \leq S(\boldsymbol{u}_a) - S(\boldsymbol{u}_b) \quad (3)$$

Theorem 1. shows that if the learned $S$ satisfies the Lipschitz continuous condition and the large margin condition in labeled samples of $Q_l$, it can also rank the unseen anomalies and background correctly and thus achieve the transferring ability. Lipschitz continuous is a common condition and controlled by the regularization strength of the deep model. Thus, meeting the large margin condition in labeled samples is the key for guaranteeing the transferring ability.

For the deviation metric learning of ADRS, we meet the large margin condition in both pixel-level and feature-level optimization. The pixel-level loss is optimized directly for the deviation ranking metric (i.e., AUC), where the discrete zero-one loss is replaced with the designed differentiable log loss. Even a correct ranking has been obtained, the penalty exists and changes according to the margin. The feature-level loss optimizes the deviation ranking of extracted features in an equivalent way, which enlarges the distance of the hypersphere centers between the anomaly and background features while decreasing their hypersphere radiuses at the same time. Both pixel-level and feature-level losses work together to strength the large margin ranking learning.

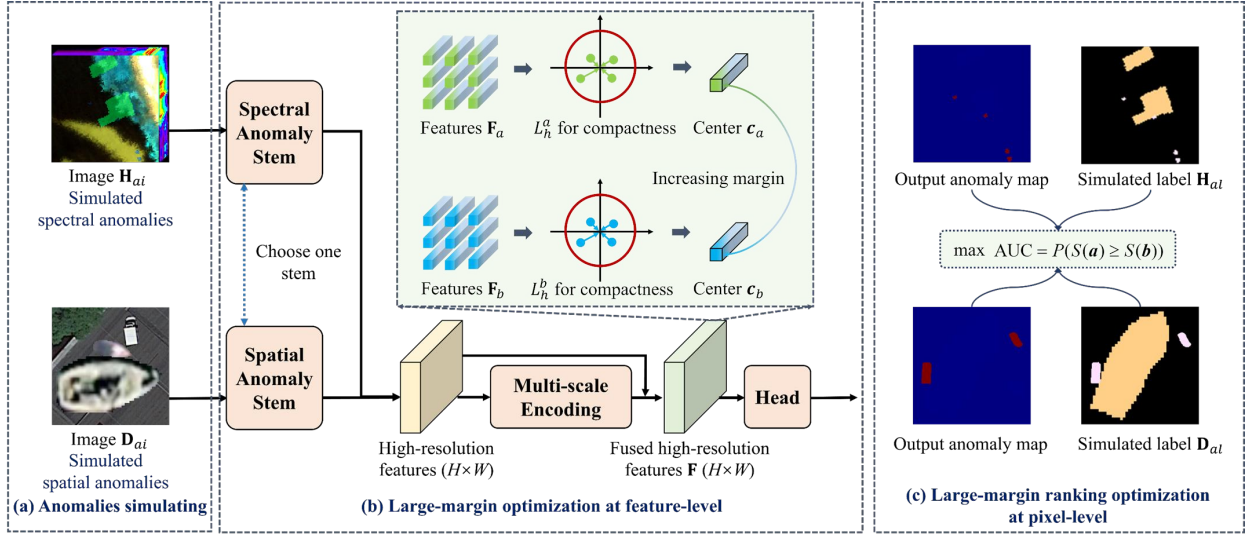

Fig. 3. The built transferring detection model with learning the deviation metric for correct ranking. According to the proven Therorem1, meeting the large-margin condition in labeled samples is the key to ensure the transferring ability of the learned deviation metric. To learn the large margin ranking, we design the pixel-level and feature-level optimization, respectively. Optimization at pixel-level (c) optimize the ranking metric AUC directly, where the discrete zero-one loss is replaced by the designed surrogate loss to be differentiable and large margin (Section 3.3). Optimization at feature-level (b) aims to enlarge the ranking margin of features, which decreases the hypersphere radiuses enclosing the anomaly and background features and also increases their center distance (Section 3.3). Besides, since the real anomalies are difficult to acquire, we simulate both spectral and spatial anomalies (a) to compute the large-margin losses.

Besides, since ADRS task is unsupervised without real anomalies, we propose an anomaly generating strategy to generate the paired labeled samples by simulating the deviation ranking relationship. The simulated samples convert ADRS from the unsupervised learning setting into the pseudo supervised setting, which are used to compute the pixel-level and feature-level ranking margin losses.

Optimized with the large margin losses (Section 3.3) and the simulated anomaly samples (detailed in Section 3.4), a transferring model for ADRS task can be built as in Fig. 3, which can be trained only once and transferred to unseen images of different modality directly.

### 3.3 Large-Margin Ranking Losses

Traditional ranking learning adopts proxy losses (e.g., cross entropy loss for the classification task [75]) or the discriminate ranking losses (e.g., the average precision (AP) loss) [76]. To be more consistent with the common ranking metric, we derive the pixel-level large-margin loss from the AUC directly and also design the feature-level loss to strengthen the large margin ranking learning.

**Pixel-level ranking loss.** We derive the loss from the AUC metric to keep the optimization process and the ranking evaluation consistent. AUC measures the probability of that $a \in \mathbf{A}$ will rank higher than $b \in \mathbf{B}$ and can be written in the integral form as in Eqs. (4) and (5),

$$\text{AUC} = P(S(\boldsymbol{a}) \geq S(\boldsymbol{b})) = \int_0^1 \text{TPR}@\text{FPR}_\eta(S) d\eta \quad (4)$$

$$\text{TPR}@\text{FPR}_\eta(S) = \max \text{TPR}(S,t) \quad s.t. \ \text{FPR}(S,t) \leq \eta \quad (5)$$

where TPR is the true positive rate and the FPR is the false positive rate. The anomaly is considered as the positive class while the background is negative. The changing false alarm rate $\eta$ is decided by the corresponding threshold $t$, which transforms the continuous anomaly map into a binary map. The TPR and FPR can be expressed with the zero-one loss $L_{01}$ (i.e., 0 for correct prediction and 1 for wrong prediction) as in Eqs. (6) and (7).

$$\text{TPR}(S,t) = \frac{\sum_{\boldsymbol{a} \in \mathbf{A}} 1 - L_{01}(S(\boldsymbol{a}),t)}{|\mathbf{A}|} \quad (6)$$

$$\text{FPR}(S,t) = \frac{\sum_{\boldsymbol{b} \in \mathbf{B}} 1 - L_{01}(S(\boldsymbol{b}),t)}{|\mathbf{B}|} \quad (7)$$

For the large-margin optimization, the sigmoid loss or the $p$-order hinge loss [77] can be used as the surrogate loss to make the discreate $L_{01}$ differentiable. However, they always need the another hyperparameter to control the margin. To tackle this, we choose to achieve the margin optimization with the help of log curve rather than the hyperparameter. The proposed surrogate loss $\overline{L}(\boldsymbol{x},t)$ for $L_{01}$ is defined in Eq. (8), which covers the four situations.

$$\overline{L}(\boldsymbol{x},t) = \begin{cases} -\log(S(\boldsymbol{x})) & \text{if } \boldsymbol{x} \in \mathbf{A} \text{ and } S(\boldsymbol{x}) \geq t \\ \dfrac{\log(S(\boldsymbol{x}))}{\log(t)} & \text{if } \boldsymbol{x} \in \mathbf{A} \text{ and } S(\boldsymbol{x}) < t \\ -\log(1-S(\boldsymbol{x})) & \text{if } \boldsymbol{x} \in \mathbf{B} \text{ and } S(\boldsymbol{x}) < t \\ \dfrac{\log(1-S(\boldsymbol{x}))}{\log(t)} & \text{if } \boldsymbol{x} \in \mathbf{B} \text{ and } S(\boldsymbol{x}) \geq t \end{cases} \quad (8)$$

For the first and third situations, although the model has already scored $\mathbf{A}$ and $\mathbf{B}$ correctly given the threshold $t$ (i.e., $S(\boldsymbol{a}) \geq t$ or $S(\boldsymbol{b}) < t$), the loss exists and encourages the larger

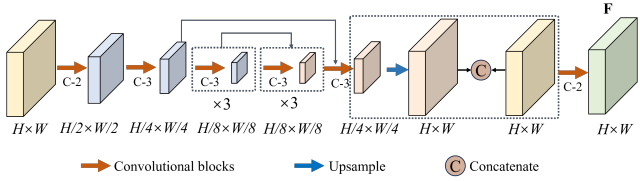

$H \times W$  $H/2 \times W/2$  $H/4 \times W/4$  $H/8 \times W/8$  $H/8 \times W/8$  $H/4 \times W/4$  $H \times W$  $H \times W$  $H \times W$

→ Convolutional blocks    → Upsample    Ⓒ Concatenate

Fig. 4. The detailed architecture of multi-scale encoding process in Fig. 3(b). The usage feature color and size are consistent with Fig. 3(b), and C-2 represents two cascaded convolution layers. The output fusing features $\mathbf{F}$ have the same spatial resolution ($H \times W$) with input image and is used to compute the high-resolution deviation score map in the convolutional

score margin. The smaller margin implies larger loss and the correlation is controlled by the log curve. If the model has given a wrong score ranking, the corresponding loss relies on both the score value and the degree of the threshold $t$. For example, when the $t$ is very large near one, a lot of anomaly pixels would be classified wrongly as the background, resulting a large and unreasonable loss. To deal with this problem, we multiply the loss with a factor $1/\log(t)$, which gives less weight to unreasonable thresholds. Replacing the $L_{01}$ in Eqs. (6) and (7) with $\overline{L}(\mathbf{x},t)$, we can get the surrogate ones denoted as $\overline{\text{TPR}}(S,t)$ and $\overline{\text{FPR}}(S,t)$ respectively as in Eq. (9).

$$\overline{\text{TPR} @ \text{FPR}_\eta}(S) = \max \overline{\text{TPR}}(S,t) \quad s.t. \quad \overline{\text{FPR}}(S,t) \leq \eta \quad (9)$$

**Theorem 2.** The surrogate $\overline{\text{TPR} @ \text{FPR}_\eta}(S)$ is a lower bound for the original $\text{TPR} @ \text{FPR}_\eta(S)$.

**Proof.** Considering Eqs. (6) and (8), $L_{01}(S(\mathbf{x}) \geq t)$ is 0 but $0 \leq \overline{L}(\mathbf{x},t) < 1$ when $\mathbf{x} \in \mathbf{A}$ and $S(\mathbf{x}) \geq t$. $L_{01}(S(\mathbf{x}) \geq t)$ is 1 but $\overline{L}(\mathbf{x},t) > 1$ when $\mathbf{x} \in \mathbf{A}$ and $S(\mathbf{x}) < t$. Thus, $\overline{\text{TPR}}(S,t)$ with the $\overline{L}$ is the lower bound of the $\text{TPR}(S,t)$ with the $L_{01}$. Similarly, $\overline{\text{FPR}}(S,t)$ is the upper bound of the original $\text{FPR}(S,t)$ considering Eqs. (7) and (8) together. Therefore, $\overline{\text{TPR}}(S,t) \leq \text{TPR}(S,t)$ and $\overline{\text{FPR}}(S,t) \geq \text{FPR}(S,t)$, and the Theorem is proved.

Theorem 2 proves the surrogate rationality of the designed differentiable large-margin $\overline{L}(\mathbf{x},t)$ for the discreate $L_{01}$. After replacing the $\text{TPR} @ \text{FPR}_\eta(S)$ in Eq. (4) with $\overline{\text{TPR} @ \text{FPR}_\eta}(S)$, we can use the Lagrange multiplier $\lambda$ to deal with the constraint of $\overline{\text{FPR}}(S,t)$ and then approximate the integral in with a discrete sum over the anchor values.

The final obtained large-margin ranking loss at pixel-level $L_p$ is given in Eq. (10), where $k$ anchors exist in the range [0,1], with each anchor corresponding to the false alarm rate $\eta_i$, threshold $t_i$, and multiplier $\Delta_i$. $\Delta_i = \eta_i - \eta_{i-1}$ for $\forall i = 1\ldots k$.

$$L_p = \min_{S,t_1,\ldots,t_k} \max_{\lambda_1,\ldots,\lambda_k} \sum_{i=1}^{k} \Delta_i (1 - \overline{\text{TPR}}(S,t_i)) \\ + \lambda_i (\overline{\text{FPR}}(S,t_i) - \eta_i \mid \mathbf{A} \mid) \quad (10)$$

**Feature-level ranking loss.** Since the remote sensing anomalies are always tiny objects, the high-resolution features are essential for preventing the loss of details. As in Fig. 3(b), two separate stems are designed to process the spectral and spatial anomalies respectively first. Both stems consist of two cascaded convolution layers, where spectral stem uses kernel size $1 \times 1$ covering spectral dimension only and spatial stem uses $3 \times 3$ covering both spatial and spectral dimensions. The output high resolution features are then processed by multi-scale blocks (as in Fig. 4), where a maximum downsampling rate of $8\times$ is set to filter out small anomaly objects. Concatenating the output context features with the previous high-resolution ones from stems, fused $\mathbf{F} \in R^{H \times W \times L}$ can be obtained, which provides both pixel-level and context-level information for each object and helps computing the deviation score with convolutional head.

To strengthen the large-margin ranking, our feature-level optimization is conducted on the multi-scale fused features $\mathbf{F} \in R^{H \times W \times L}$ with the original image spatial size $H \times W$ and the feature dimension $L$. The anomaly features $\mathbf{F}_a$ and background features $\mathbf{F}_b$ are extracted from $\mathbf{F}$ according to the sample label. Specifically, we decrease the hypersphere radiuses enclosing the anomaly and background features and also increase their center margin.

Given $\mathbf{F}_a$, its hypersphere center $\mathbf{c}_a \in R^L$ can be computed as the mean value along the spatial dimension as in Eq. (11).

$$\mathbf{c}_a = \text{mean } \mathbf{f}_a, \quad \mathbf{f}_a \in \mathbf{F}_a \quad (11)$$

To decrease the radius $R_a$ in Eq. (12) while making the hypersphere including $\mathbf{F}_a$ as much as possible, the hypersphere optimization $L_h^a$ for $\mathbf{F}_a$ is formulated as the minimization problem in Eq. (13). The optimization $L_h^b$ for $\mathbf{F}_b$ can be obtained in the similar way.

$$R_a^2 = \underset{\mathbf{f}_a \in \mathbf{F}_a}{\text{mean}}(\|\mathbf{f}_a - \mathbf{c}_a\|^2) \quad (12)$$

$$L_h^a = R_a^2 + \beta \underset{\mathbf{f}_a \in \mathbf{F}_a}{\text{mean}}(\max\{\|\mathbf{f}_a - \mathbf{c}_a\|^2 - R_a^2, 0\}) \quad (13)$$

$L_h^a$ and $L_h^b$ make the corresponding hyperspheres compact and the hypersphere centers can represent the overall feature distribution in a certain extent. With the constraints of $L_h^a$ and $L_h^b$, the feature level loss $L_f$ enlarges the ranking margin of $\mathbf{F}_a$ and $\mathbf{F}_b$ by increasing the distance of the anomaly hypersphere center $\mathbf{c}_a$ and the background hypersphere center $\mathbf{c}_b$. The $L_f$ is formulated in Eq. (14). Since the three terms have the same order of magnitude and importance, the loss ratio is set $1:1:1$.

$$L_f = -\|\mathbf{c}_b - \mathbf{c}_a\|^2 + L_h^a + L_h^b \quad (14)$$

In total, the pixel-level loss $L_p$ and the feature-level loss $L_f$ work together for the large-margin score ranking target as in Eq. (15) where the $w$ controls the balance.

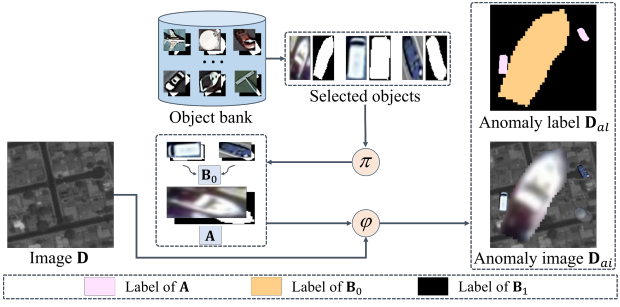

Fig. 5. The designed workflow for the spatial anomaly simulation with high spatial resolution images as input. We built an additional object bank with over 650000 instances, where the objects from different images are randomly selected and resized to preset area ranges to simulate the deviating ranking relationship of $\mathbf{A}$, $\mathbf{B}_0$ and $\mathbf{B}_1$.

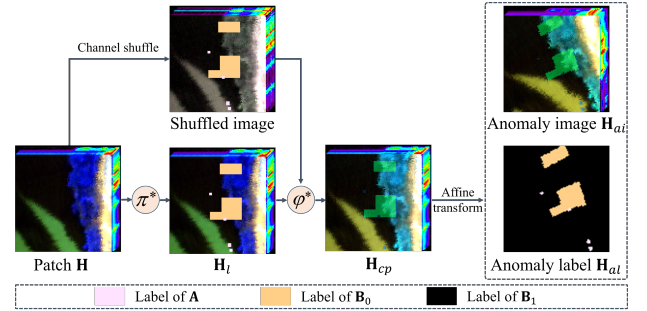

Fig. 6. The designed workflow for the spectral anomaly simulation with several hyperspectral benchmarks. We use channel shuffle operation to create the spectral deviation relationship, and the simulated anomalies have smaller size than spatial anomalies to better align with the practical situation.

$$L = L_p + wL_f \qquad (15)$$

### 3.4 Anomaly Sample Simulation

Since the ADRS task is unsupervised while the large margin condition mentioned above needs to be satisfied in labeled samples, we propose the simulation strategy to generate the paired anomaly samples. To simulate samples covering all the remote sensing modalities, we simulate both the anomalies in spectral domain and in spatial domain. Spectral anomalies deviate from the surroundings with properties in both spectral and spatial aspects (e.g., the hyperspectral modality) while the spatial anomalies deviate in the spatial properties only for the modality with few channels (e.g., SAR) [4].

For the simulation of the spatial anomalies, the workflow is designed with the large-scale iSAID dataset to provide the rich spatial details. Fig. 5 shows the overall workflow. Since the anomalies are always small in size, we simulate the large size objects $\mathbf{B}_0$ of background $\mathbf{B}$ explicitly in addition to $\mathbf{A}$ to train the model being aware of the object size. Thus, $\mathbf{B} = \mathbf{B}_0 + \mathbf{B}_1$, where $\mathbf{B}_1$ is the remaining background part. The input image $\mathbf{D}$ is randomly selected from the iSAID dataset, serving as the background $\mathbf{B}_1$. The anomaly tiny objects $\mathbf{A}$ and large size objects $\mathbf{B}_0$ are both selected from the pre-built object bank, which includes the 650,000 instances from the iSAID dataset. The $\pi$ operation separates the selected objects into two groups ($\mathbf{A}$ and $\mathbf{B}_0$) and resizes them into the preset range (Generally, the size of $\mathbf{B}_0$ is obviously larger than $\mathbf{A}$). Since $\mathbf{A}$ and $\mathbf{B}_0$ originally do not belong to the $\mathbf{D}$ and the $\mathbf{B}_0$ has an obviously larger area than $\mathbf{A}$, the desired ranking inequality $S(\boldsymbol{a} \in \mathbf{A}) > S(\boldsymbol{b}_0 \in \mathbf{B}_0) > S(\boldsymbol{b}_1 \in \mathbf{B}_1)$ can be assumed true. Finally, the $\varphi$ operation pastes the resized $\mathbf{A}$ and $\mathbf{B}_0$ into $\mathbf{D}$, and obtains the anomaly image $\mathbf{D}_{ai}$. The corresponding label $\mathbf{D}_{al}$ can also be obtained.

For the simulation of the spectral anomalies ($\mathbf{H}_{ai}$, $\mathbf{H}_{al}$), we inherit the main workflow from the prior TDD model [27], where the data argumentation technique of channel shuffling is used to create the spectral deviation of anomalies. Fig. 6 shows the simulation workflow. Given input hyperspectral patch $\mathbf{H}$, $\pi^*$ operation first randomly selects locations and obtains $\mathbf{H}_l$ for generating $\mathbf{A}$ and $\mathbf{B}_0$ according to the preset area range. The selected locations in $\mathbf{H}_l$ are then replaced by the corresponding spectra in shuffled images (i.e., $\varphi^*$ operation). Since $\mathbf{A}$ and $\mathbf{B}_0$ are violently shuffled in spectral dimension and $\mathbf{B}_0$ has a larger area than $\mathbf{A}$, the ranking $S(\boldsymbol{a} \in \mathbf{A}) > S(\boldsymbol{b}_0 \in \mathbf{B}_0) > S(\boldsymbol{b}_1 \in \mathbf{B}_1)$ can be assumed to be true in output $\mathbf{H}_{cp}$ similar to the spatial anomaly simulation. To increase the shape diversity, affine transformation is finally conducted to output the ($\mathbf{H}_{ai}$, $\mathbf{H}_{al}$). Three hyperspectral benchmarks (WHU-Hi-LongKou, WHU-Hi-HanChuan, and WHU-Hi-HongHu) [78] are used to provide the input of the simulation workflow.

In total, the simulated anomaly samples can make the learned $S$ be optimized with the proposed large-margin losses (Section 3.3). According to the theorems proved in Section 3.2, once the trained $S$ has achieved the large-margin performance in the simulated samples, it can also detect the unseen images of the different modalities and keep the deviation inequality hold.

To show the overall workflow of proposed model, we have provided a pseudo code in Algorithm 1 including both training and testing stages.

---

**Algorithm 1** UniADRS

**Training stage (1 iteration):**
1: Simulate one paired spectral anomaly ($\mathbf{H}_{ai}$, $\mathbf{H}_{al}$)
2: Set spectral stem for $\mathbf{H}_{ai}$
3: Forward computation and output one anomaly map
4: Simulate one paired spatial anomaly ($\mathbf{D}_{ai}$, $\mathbf{D}_{al}$)
5: Set spatial stem for $\mathbf{D}_{ai}$
6: Forward computation and output another anomaly map
7: Compute loss $L_p$ and $L_f$ for both $\mathbf{H}_{ai}$ and $\mathbf{D}_{ai}$
8: Network backward

**Testing stage for any unseen image:**
1: Set spectral stem for hyperspectral modality or spatial stem for visible light, SAR, infrared and low-light modalities
2: Forward computation and output one anomaly map

**Model Input:** One image of any remote sensing modality and scene
**Model Output:** One corresponding anomaly map

4## 4. EXPERIMENTAL RESULTS

### 4.1 Experimental Settings

In this section, we describe how the proposed transferring model was validated in five modalities, i.e., hyperspectral, visible light, SAR, infrared, and low-light, to show its cross-modal ability. The proposed model is named as the **uni**fied **a**nomaly **d**etector in **r**emote **s**ensing (UniADRS). In this section, the comparative experiments are firstly described with other non-transferring models, which were trained separately on each scene. Then, the model analysis results and the model efficiency are also discussed.

#### 4.1.1 Constructed Multi-Modal Dataset

We built a multi-modal dataset for the ADRS task, with hyperspectral, visible light, SAR, infrared, and low-light modalities (as detailed in Table 1). The images in the dataset cover various scenes, sensor types, and resolutions. All the test images of five modalities were unseen at test stage to verify the detector transferability. The 82 hyperspectral scenes were collected from the Cri dataset [29] and the two large-scale unmanned aerial vehicle (UAV)-borne datasets of WHU-Hi-Park and WHU-Hi-Station [31]. For the low-light modality, we first captured 50 scenes at night and then doubled this by data augmentation to make the overall size balanced. The multi-modal dataset will be made publicly available.

#### 4.1.2 Comparison Methods and Evaluation Metrics

Due to the property of the high spectral resolution, the hyperspectral modality has many unique models and was considered separately from the other modalities.

The comparative models for the hyperspectral modality were the global RX detector (GRX) [15], the abundance- and dictionary-based low-rank decomposition (ADLR) detector [79], the collaborative representation based (CRD) detector [20], the spectral constraint autoencoder (SC_AAE) detector [23], the deep low-rank prior based detector (DeepLR) [31], and the TDD method [27]. The comparison methods cover the three categories of statistical-based, representation-based, and deep learning based methods.

The comparative models for the remaining four modalities were GRX [51], a convolutional autoencoder (CAE) [80], a variational autoencoder (VAE) [81], the saliency-based method proposed by Cai *et al* [41] and an adversarial autoencoder (AAE) [21]. The implementation of these methods was adapted from the related ADRS studies [21], [41], [51], [81]. Besides, we also compared UniADRS with the state-of-art industrial anomaly detection model UniAD [35]. To adapt the UniAD for the small objects in ADRS task, the input size is increased from 224 to 1024. The remaining settings are kept same as [35].

The detection performance is evaluated with multi-parameter 3D receiver operating characteristic (3D ROC) curves [82]. Compared to 2D ROC curves, the threshold dimension is additionally considered and can provide more comprehensive information. The used metrics are the widely used $AUC_{(D,F)}$, the target detectability $AUC_{TD}$, the background suppressibility $AUC_{BS}$, and the overall detection probability $AUC_{ODP}$. Each metric value is positively correlated with the detection performance.

#### 4.1.3 Implementation Details

The hyperparameters of the comparative hyperspectral models are set following [27]. The CAE architecture was Unet with a ResNet50 backbone. For the SAR modality, speckle removal was conducted before applying the AAE method, following [21]. When simulating the spectral anomalies, we controlled the **A** area in ratio range [0.0064, 0.0225] and $\mathbf{B}_0$ in range [0.0225, 0.5]. Similarly, we controlled the **A** in the range [0.02, 0.06] and $\mathbf{B}_0$ in range [0.06, 0.5] for simulated spatial anomalies. The feature-level optimization loss and the pixel-level loss were added at a ratio of $w$=0.1. The UniADRS was optimized with the Adam optimizer (learning rate 0.01, weight decay 1e−5, batch size 1) over 100 epochs.

At test stage, we use the trained UniADRS on unseen images without any further fine-tuning. Spectral stem is used for hyperspectral modality and spatial stem for visible light, SAR, infrared and low-light modalities. We use the channel processing technique from [27] to deal with the varying channels of hyperspectral modality. Overlap technique is also used to improve the performance [27], where we set patch size 50 for hyperspectral modality and 100 for other modalities. The sensitivity analysis about the inferring patch size is reported in Section 4.3.5. The CPU was an Intel(R) Xeon(R) Gold 5218R CPU @ 2.10 GHz with 251 GB memory, and the GPU was a NVIDIA GeForce RTX 4090 with 24 GB memory.

### 4.2 Comparison Results

In all the five modalities, proposed UniADRS inferred the test images directly while the comparative models were retrained for each image. The quantitative results are reported in Table 2 and Table 3. Fig. 7-11 visualizes the obtained anomaly maps on five modalities, respectively.

*Hyperspectral modality.* In Table 2, UniADRS is the only model that achieves an $AUC_{(D,F)}$ metric score of higher than 0.97 and an $AUC_{ODP}$ metric score of higher than 1.35 on all three datasets (82 hyperspectral scenes). Although the TDD model shows satisfactory transferability on the Cri dataset, the metric scores drop dramatically on the UAV-borne WHU-Hi Park and Station datasets (AUC$_{(D,F)}$ 0.67 and 0.71, respectively). Despite the tiny anomaly sizes (especially the second example in Fig. 7), the obtained anomaly map of UniADRS has the best discriminability.

TABLE 1
The Detailed Information of Constructed Multi-modal Dataset for the ADRS Task

| Modality | Source | Spatial resolution | Image size | Scene number | Anomalies |
|---|---|---|---|---|---|
| Hyperspectral | Nuance Cri; Nano-Hyperspec | 4–8 cm/pixel | 400×400; 200×200 | 82 | Plastic plane, metal object, etc. [31] |
| Visible light | Google Earth | 0.5–2 m/pixel | 1044×915 | 100 | Military camouflage [48], aircraft [12] |
| SAR | Gaofen-3; Sentinel-1 | 3–10 m/pixel | 256×256 | 100 | Various ships [10], [57] |
| Infrared | \ | \ | 173×98; 407×305 | 100 | Car, dim lamp, etc. [37] |
| Low-light | Indigo NV-400-M | \ | 2048×2048 | 100 | Toy car, plane, tank, etc. |

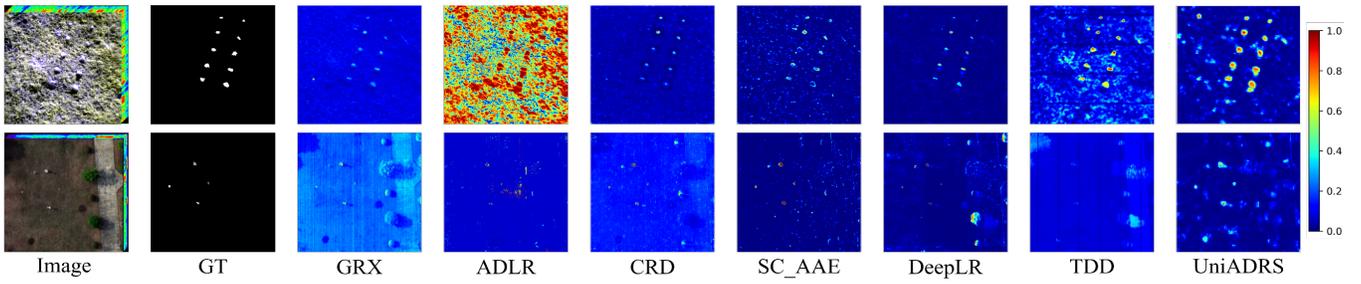

Fig. 7. Typical anomaly detection results for the hyperspectral modality, where the anomalies include rocks (first row), fabric camouflage objects (second row) and metal objects (second row).

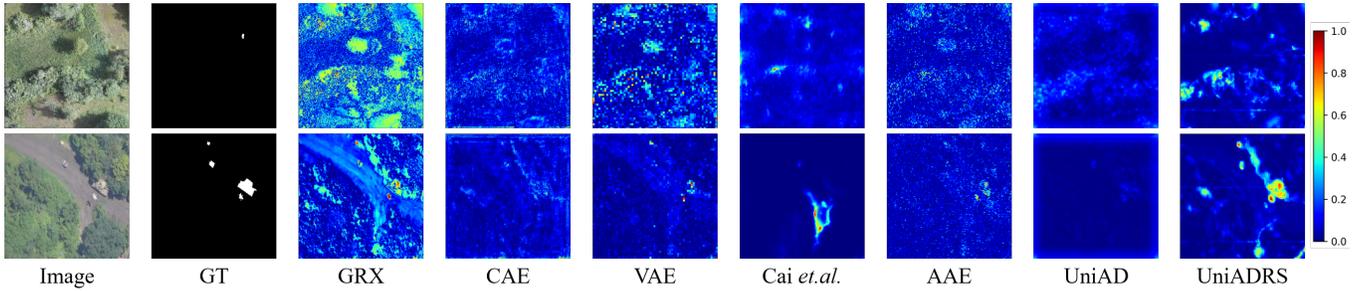

Fig. 8. Typical anomaly detection results for the visible light modality, where the anomalies include the camouflage net (first row), a tank and a drone (second row).

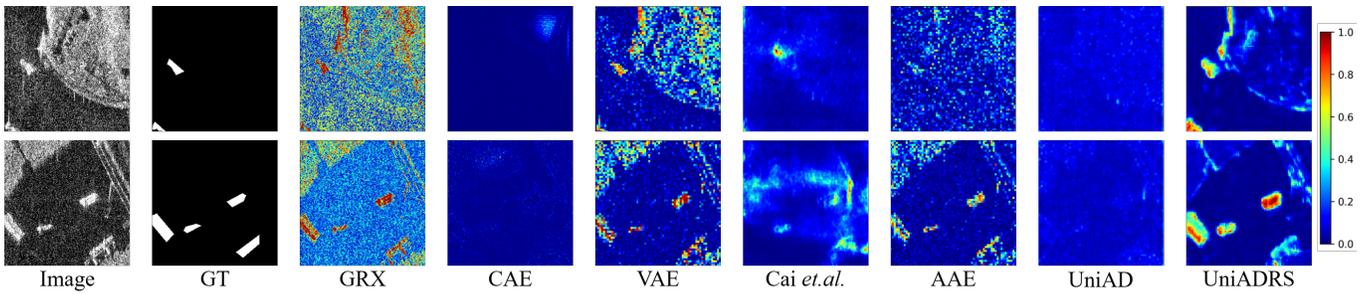

Fig. 9. Typical anomaly detection results for the SAR modality, where the anomalies include various ships.

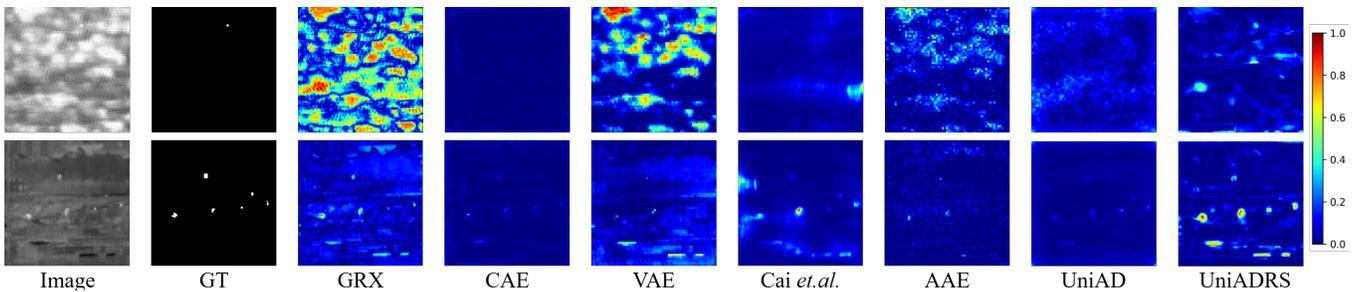

Fig. 10. Typical anomaly detection results for the infrared modality, where the anomalies include the plane (first row) and peoples (second row).

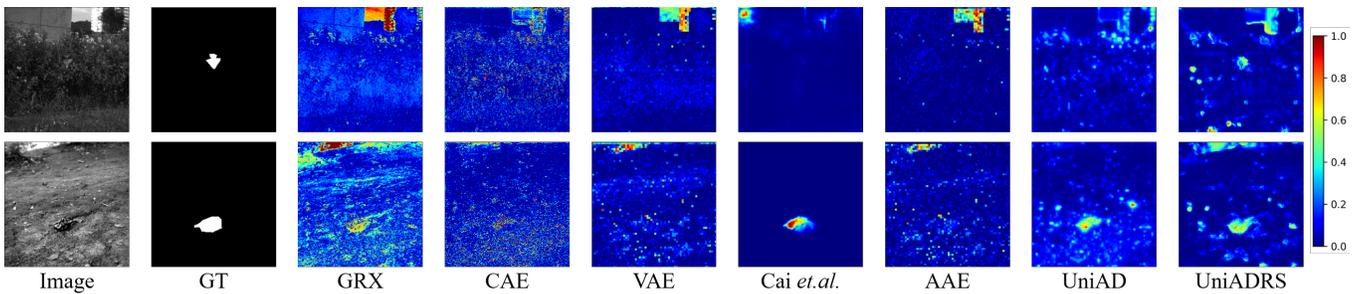

Fig. 11. Typical anomaly detection results for the low-light modality, where the anomalies include a toy plane (first row) a toy tank (second row).

*Visible light modality.* Table 3(a) reports the related results. UniADRS achieves the best performance under the AUC$_{(D,F)}$ and AUC$_{BS}$ metrics. Proposed model and UniAD are the only two models with an AUC$_{(D,F)}$ score of higher than 0.80. Although the AUC$_{TD}$ score of our model is lower than that of GRX, this could be improved with a simple post-processing of image stretching. In Fig. 8, the first sample is inconspicuous and many detectors fail to find it. The second scene comes from the Russo-Ukrainian War, where a Russia tank is hiding and a Ukrainian UAV attempted to blow up. Many of the methods correctly locate the anomalies in this scene, but with an incomplete shape. In contrast, UniADRS achieves the best tradeoff between detection completeness and false alarms.

*SAR modality.* Table 3(b) reports the related results. The detection on SAR modality is relatively easy for most of the SAR scenes, because the anomalies (i.e., ships) lie in a homogeneous sea background. Most models can achieve the AUC$_{(D,F)}$ higher than 0.85. Similar to the visible light modality, proposed model surpasses the second-place UniAD by around 4 points on AUC$_{(D,F)}$ metric. For the examples in Fig. 9, many models fail to process the speckle noise and the obtained anomaly maps are full of salt-and-pepper noise such as the GRX, VAE and AAE. Since our large-margin learning has seen many spatial anomalies and learned the context modeling ability, proposed model can suppress most noises successfully.

*Infrared modality.* Table 3(c) reports the related results. Proposed model achieves the highest AUC$_{(D,F)}$ score of 0.94, which surpasses the supervised result 0.91 in [39], even when inferred directly. In Fig. 10, the anomalies in first example are extremely tiny and many model fails to detect it. The second example has 6 anomalies in total. In manual interpretation, only 2-3 anomalies can be seen in many comparative anomaly maps while our anomaly map can find 5 anomalies easily.

*Low-light modality.* Table 3(d) reports the related results. The captured low-light dataset seems more challenging than the remaining modalities due to the night environment. Many models achieve AUC$_{(D,F)}$ lower than 0.70 while our model can still get the optimal result 0.84, showing a robust transferring ability. Due to the camouflage property of given examples in

TABLE 2
Quantitative Results for the Hyperspectral Modality, Dozens of Scenes in WHU-Hi Park and WHU-Hi Station are Evaluated Together

| Method | **AUC$_{(D,F)}$** | AUC$_{TD}$ | AUC$_{BS}$ | AUC$_{ODP}$ | **AUC$_{(D,F)}$** | AUC$_{TD}$ | AUC$_{BS}$ | AUC$_{ODP}$ |
|---|---|---|---|---|---|---|---|---|
| | \multicolumn{4}{c}{Cri (1 scene)} | | | | | |
| GRX | 0.9678 | 1.1932 | 0.8782 | 1.1036 | 0.9379 | 1.3091 | 0.8099 | 1.2432 |
| ADLR | 0.9579 | **1.9253** | 0.3159 | 1.2833 | 0.8234 | 1.0784 | 0.7995 | 1.2311 |
| CRD | 0.9186 | 1.1350 | 0.8738 | 1.0902 | 0.9095 | 1.1046 | 0.8514 | 1.1370 |
| SC_AAE | 0.8849 | 1.1355 | 0.8608 | 1.1114 | 0.9579 | 1.1798 | 0.9509 | 1.2149 |
| DeepLR | 0.9815 | 1.2465 | **0.9687** | 1.2337 | 0.9736 | 1.1837 | **0.9607** | 1.1972 |
| TDD | 0.9915 | 1.6298 | 0.8793 | 1.5176 | 0.6712 | 0.7391 | 0.0464 | 0.7855 |
| UniADRS | **0.9970** | 1.6755 | 0.9472 | **1.6257** | **0.9748** | 1.4181 | 0.9331 | **1.3764** |
| | WHU-Hi Station (54 scenes) | | | | Average | | | |
| GRX | 0.8988 | 1.1441 | 0.8163 | 1.1628 | 0.9348 | 1.1304 | 0.8547 | 1.1146 |
| ADLR | 0.9260 | 1.3566 | 0.8650 | **1.3696** | 0.9024 | 1.0861 | 0.8293 | 1.1414 |
| CRD | 0.9722 | 0.9763 | 0.9719 | 1.0038 | 0.9334 | 0.9968 | 0.9109 | 1.0168 |
| SC_AAE | 0.9708 | 1.0455 | 0.9701 | 1.0740 | 0.9379 | 1.0611 | 0.9596 | 1.0823 |
| DeepLR | 0.9853 | 1.1169 | **0.9825** | 1.1288 | 0.9801 | 1.0914 | **0.9723** | 1.0999 |
| TDD | 0.7190 | 0.2051 | 0.5940 | 0.7991 | 0.7939 | 0.5385 | 0.4372 | 0.7519 |
| UniADRS | **0.9860** | **1.3896** | 0.9512 | 1.3548 | **0.9859** | **1.2608** | 0.9530 | **1.2353** |

TABLE 3
Quantitative Results for the Visible Light, SAR, Infrared, and Low-Light Modalities

| Method | **AUC$_{(D,F)}$** | AUC$_{TD}$ | AUC$_{BS}$ | AUC$_{ODP}$ | **AUC$_{(D,F)}$** | AUC$_{TD}$ | AUC$_{BS}$ | AUC$_{ODP}$ |
|---|---|---|---|---|---|---|---|---|
| | (a) Visible light modality | | | | (b) SAR modality | | | |
| GRX | 0.7292 | **1.1506** | 0.5210 | 0.9425 | 0.8938 | **1.5250** | 0.7931 | **1.4243** |
| CAE | 0.7970 | 0.8771 | 0.7715 | 0.8516 | 0.8281 | 0.9118 | 0.8210 | 0.9047 |
| VAE | 0.6891 | 1.0159 | 0.5552 | 0.8819 | 0.8816 | 1.3315 | 0.8495 | 1.2995 |
| Cai *et al.* | 0.7567 | 0.9205 | 0.7005 | 0.8644 | 0.8610 | 1.0612 | 0.8347 | 1.0349 |
| AAE | 0.7101 | 0.9260 | 0.6375 | 0.8534 | 0.8831 | 0.9699 | 0.8757 | 0.9626 |
| UniAD | 0.8546 | 1.0217 | 0.7931 | **0.9603** | 0.9102 | 1.0678 | 0.8329 | 0.9905 |
| UniADRS | **0.8948** | 0.9207 | **0.8901** | 0.9160 | **0.9595** | 0.9959 | **0.9549** | 0.9913 |
| | (c) Infrared modality | | | | (d) Low-light modality | | | |
| GRX | 0.6814 | 1.0899 | 0.4543 | 0.8629 | 0.6684 | **1.0900** | 0.4647 | **0.8863** |
| CAE | 0.8291 | 0.9297 | 0.8180 | 0.9187 | 0.6246 | 0.6620 | 0.6005 | 0.6380 |
| VAE | 0.7301 | **1.2339** | 0.4902 | 0.9941 | 0.5703 | 0.7299 | 0.4899 | 0.6495 |
| Cai *et al.* | 0.8853 | 1.2242 | 0.8415 | **1.1805** | 0.8248 | 0.9900 | 0.8049 | 0.9701 |
| AAE | 0.7557 | 1.0686 | 0.6598 | 0.9727 | 0.6694 | 0.8224 | 0.6196 | 0.7726 |
| UniAD | 0.8348 | 0.9145 | 0.8054 | 0.8850 | 0.7716 | 0.8563 | 0.7343 | 0.8191 |
| UniADRS | **0.9437** | 0.9820 | **0.9394** | 0.9778 | **0.8336** | 0.8558 | **0.8291** | 0.8513 |

Fig. 11, proposed model is the only to locate the anomaly with discriminative boundary and high confidence.

### 4.3 Model Analysis
#### 4.3.1 Ablation of the Model Optimization

Pixel-level and feature-level optimization are proposed for the large-margin deviation ranking target. To show the superiority, we compared it with prior ranking losses (proxy cross-entropy [75] and the average precision ranking [76]), large margin losses (sigmoid and hinge losses) [77] and the proposed pixel-level loss only. We integrate the large-margin losses into our differentiable AUC framework for fair comparison. Table 4 reports the related results. The results with different large-margin surrogate losses show better performance than the average precision ranking, which are consistent with our proven Theorem 1. Although cross-entropy loss is designed originally for the classification task, it has shown strong robustness for our deviation ranking task. Benefiting from considering both the ranking margin and the rationality of the threshold, proposed pixel-level loss has achieved the best transferring performance than the prior ranking losses and large-margin losses. Optimizing the model with pixel-level and feature-level losses together, the average $AUC_{(D,F)}$ performance is promoted further from 0.9293 to 0.9416.

#### 4.3.2 Statistics of the Cover Radius δ

As proven in the Theorem. 1, the cover radius $\delta$ measures the difference of simulated labeled samples and the unseen samples at the test stage, which is positively related with the demanded lowest margin in labeled samples for the transferring ability. The quantitative results in Section 4.2 have already shown the model meets the lowest margin demand in simulated samples and achieve transferring ability. We analyze the $\delta$ further in this section to show the learned representative distance of different modalities.

For each pixel of unseen images, its $\delta$ is the smallest radius with the same kind of pixels (anomaly or background) in simulated samples. For each test modality, the modality $\delta$ is treated as the max $\delta$ of all the pixels (defined in Theorem 1). The radius is computed with the corresponding Euclidean distance in the feature space of **F** (defined in Section 3.3). We report the resulting $\delta$ with different number of simulated images (20, 40, 60, 80), and each result is repeated four times to compute the mean (represented in broken line) and standard deviation ((represented in color block).

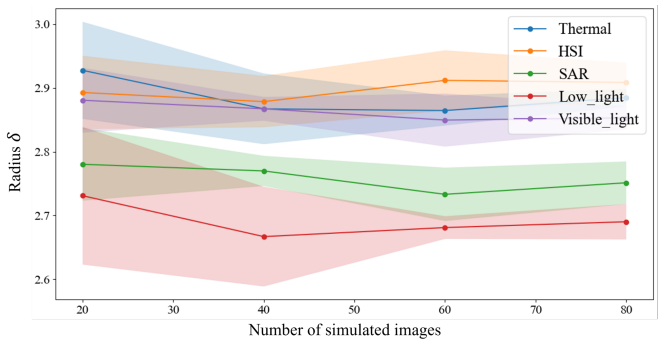

Fig. 12. The statistical radius $\delta$ between the simulated anomaly images and the unseen test images. Since the max radius value for the activated feature is in range [0, 64], the low radiuses in [2.6, 3.0] imply a low margin demand in simulated samples and the high transferring robustness.

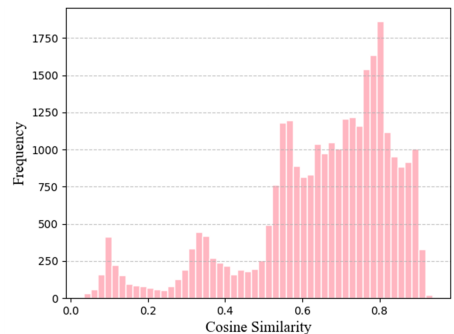

Fig. 13. The statistical cosine similarity between the simulated anomaly spectra and the background. Most spectra are in range [0.5, 0.9], which shows a weak spectra difference and high detection difficulty.

TABLE 4
Ablation Results for the Designed Model Optimization Loss

| Optimization loss | Hyperspectral $AUC_{(D,F)}$ | Visible light $AUC_{(D,F)}$ | SAR $AUC_{(D,F)}$ | Infrared $AUC_{(D,F)}$ | Low-light $AUC_{(D,F)}$ | Average $AUC_{(D,F)}$ |
|---|---|---|---|---|---|---|
| Proxy cross-entropy ranking | 0.9409 | 0.8937 | **0.9634** | 0.9296 | 0.8213 | 0.9187 |
| Average precision ranking | 0.9252 | 0.8464 | 0.8891 | 0.8755 | 0.7194 | 0.8511 |
| AUC+large margin sigmoid | 0.9533 | 0.8896 | 0.9145 | 0.8802 | 0.8507 | 0.8977 |
| AUC+large margin hinge | 0.9308 | 0.8616 | 0.9310 | 0.8318 | 0.7924 | 0.8695 |
| Pixel level | 0.9641 | 0.8851 | 0.9506 | 0.9337 | **0.8434** | 0.9293 |
| Pixel and feature level loss | **0.9859** | **0.8948** | 0.9595 | **0.9437** | 0.8336 | **0.9416** |

TABLE 5
Ablation Results for the Designed Anomaly Sample Simulation Strategy

| Spectral stem | Spatial stem | $\mathbf{B}_0$ Simulation | Hyperspectral $AUC_{(D,F)}$ | Visible light $AUC_{(D,F)}$ | SAR $AUC_{(D,F)}$ | Infrared $AUC_{(D,F)}$ | Low-light $AUC_{(D,F)}$ | Average $AUC_{(D,F)}$ |
|---|---|---|---|---|---|---|---|---|
| √ | × | × | 0.8668 | 0.7390 | 0.8674 | 0.7683 | 0.6988 | 0.8106 |
| √ | × | √ | 0.9377 | 0.8285 | 0.8038 | 0.8012 | 0.7375 | 0.8549 |
| × | √ | × | 0.9256 | 0.7916 | 0.9211 | 0.8168 | 0.8136 | 0.8743 |
| × | √ | √ | 0.9538 | 0.8597 | **0.9667** | 0.8703 | 0.7815 | 0.9056 |
| √ | √. | √ | **0.9859** | **0.8948** | 0.9595 | **0.9437** | 0.8336 | **0.9416** |

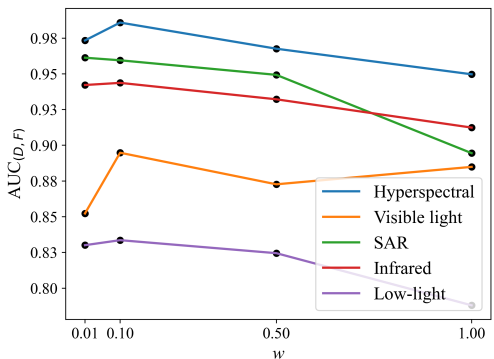

Fig. 14. The sensitivity analysis about the loss weighted parameter *w*, where most modalities achieve the best accuracy at the *w* 0.1.

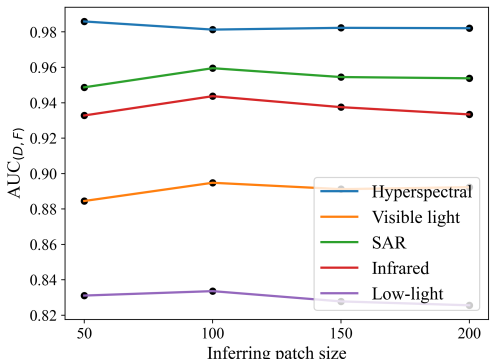

Fig. 15. The sensitivity analysis about the inferring patch size, where proposed model is robust to the changes with a maximum difference of 1 point.

In the results of all the five modalities (Fig. 12), our model has obtained similar representation between the simulated images and the unseen test images, where the radius $\delta$ is lower than 3.0. Since the feature dimension of $\mathbf{F}$ is 64 in practice and the radius range is between [0,64] after the sigmoid activation. The radius 3.0 is very small compared the max value 64. As the selected image number grows, closer representation may appear and the resulting $\delta$ decreases in many modalities (e.g., visible light and SAR). The low $\delta$ value implies the low margin demand in simulated samples and the high transferring robustness.

*4.3.3 Ablation of the Sample Simulation Strategy*

For the proposed UniADRS model, we simulate both the spectral anomalies and spatial anomalies, where the background, large normal objects, and anomalies are explicitly modeled. With the simulated samples, UniADRS can be trained with the designed large margin losses. The prior qualitative comparing results have shown that the explicit model for large normal objects $\mathbf{B}_0$ can decrease the false alarms effectively.

We conducted the ablation experiments from two aspects: whether to simulate the large normal objects $\mathbf{B}_0$ and whether to simulate both kinds of anomalies. The related results are shown in Table 5. Comparing row 1 with row 2, and row 3 with row 4, it is clear that the $\mathbf{B}_0$ simulation can bring a stable gain in most modalities especially for the hyperspectral (7 points) and the visible light (9 points) modalities. For the infrared and low-light modalities, the increase is relatively lower (around 4 points). We deduce that the gain obtained from the $\mathbf{B}_0$ simulation is positively correlated with the scene complexity. Comparing the results of using spectral anomalies or spatial anomalies only, the spatial anomaly simulation can result in a more robust performance in most modalities, regardless of the $\mathbf{B}_0$ simulation. The inclusion of the spatial and spectral stems helps the detector to better fuse both the spatial and spectral features.

*4.3.4 Difficulty of the Simulated Spectral Anomalies*

Channel shuffling operation is used to decrease the spectral correlation of simulated anomalies and the background. Generally, the higher the correlation, the greater the detection difficulty. To quantitatively analyze the sample difficulty, we use the cosine similarity to compute the correlation degree.

Fig. 13 shows the statistical results from over 10000 spectral anomalies. For each simulated anomaly spectrum, we recorded the cosine distance between it and the surrounding background. The resulting statistical distribution is not uniform, where most results lie at the range from 0.5 to 1.0 and the peak value appears around the 0.8, implying a high correlation and detection difficulty. From this perspective, the simulated spectral samples are hard examples, which helps the learned model be more robust for the unseen anomalies.

*4.3.5 Sensitivity Analyses*

UniADRS is trained to be a deviation metric with designed large-margin ranking losses, where pixel-level and feature-level losses are weighted together with *w* to supervise the model. To report the related sensitivity analysis, we varied the *w* from 0.01 to 1.0 and observed the corresponding accuracy in five test modalities. The results are reported in Fig. 14. The changing of *w* has an obvious effect on all the modalities, which can cause a maximum difference of 6 points in accuracy. It implies that feature-level and pixel-level optimizations may not be entirely consistent and require some designed reconciliation. Except for SAR, all other modalities achieve the best accuracy at the *w* 0.1, which is finally chosen as the default setting.

At the test stage, we use overlap setting to improve the performance, where each image is inferred in overlapped patches. Fig. 15 reports the related sensitivity analysis about the inferring patch size. The results show that proposed model is robust to inference sizes, causing at most a 1-point difference. From the perspective of best accuracy, the optimal size for the hyperspectral modality is 50 and 100 for other modalities.

*4.4 Model Efficiency*

One of the great advantages of the proposed UniADRS model is the elimination of training for each unseen image. In this section, the efficiency of UniADRS is investigated by computing the model processing time for each modality.

Table 6 lists the recorded processing times for the hyperspectral modality. Since the comparative models belong to transductive models and need to be trained with test images, their recorded processing times include both the training and testing stages. In contrast, proposed model can infer the unseen test modalities directly and the recorded processing time includes the testing time only. The current state-of-the-art model of DeepLR needs around 3 and 4 hours for the WHU-Hi Park and WHU-Hi Station datasets, respectively. Although TDD can deal with the WHU-Hi scenes in less than 2 min, the accuracy is not satisfactory, as shown in Table 2. Keeping the highest accuracy performance, the proposed UniADRS model can process the scenes faster than the representation-based and deep learning based methods, and the time is closer to that of GRX.

TABLE 6
Efficiency Comparison for the Hyperspectral Modality.

| Method | Cri | WHU-Hi Park | WHU-Hi Station |
|---|---|---|---|
| GRX | 3.73s | 51.91s | 71.96s |
| ADLR | 1258.50s | 25405.03s | 34227.51s |
| CRD | 1024.84s | 37427.60s | 37181.45s |
| SC_AAE | 128.91s | 9944.33s | 21728.03s |
| DeepLR | 31.49s | 10187.13s | 13714.35s |
| TDD | 4.21s | 94.51s | 166.38s |
| UniADRS | 5.14s | 59.13s | 121.98s |

TABLE 7
Efficiency Comparison for the Visible Light, SAR, Infrared, and Low-Light Modalities.

| Method | Visible light modality | SAR modality | Infrared modality | Low-light modality |
|---|---|---|---|---|
| GRX | 37.52s | 56.67s | 41.08s | 123.12s |
| CAE | 172.64s | 162.23s | 646.43s | 129.30s |
| VAE | 113.77s | 227.38s | 71.60s | 185.98s |
| Cai *et al.* | 268.73s | 371.39s | 1236.66s | 1452.26s |
| AAE | 160.63s | 153.19s | 659.41s | 1202.76s |
| UniAD | 7440.16s | 3120.53s | 5760.88s | 6324.86s |
| UniADRS | 83.68s | 107.69s | 64.64s | 61.75s |

Table 7 lists the recorded processing times for the remaining four modalities without spectral information. Proposed UniADRS model has surpassed all the comparative deep models by at least an order of magnitude, and achieved closer performance with GRX. Low-light modality is a special case, where GRX takes more time than proposed model due to its large image size (2048×2048 in Table 1). Given the same image size, GRX processes the image pixel-by-pixel with CPU while proposed model can utilize the parallel computing capability of the GPU and constitute a batch for a single forward propagation.

The obtained results fully prove the real-time performance of UniADRS, and its ability to process large-scale hyperspectral scenes in real time.

## 5. CONCLUSION

In this study, we designed a transferring anomaly detector for different remote sensing modalities by transferring the learning target from certain image distribution to the image-independent deviation metric. To guide the learning of deviation metric, we firstly theoretically prove that although the cross-modality images are unseen at training stage, once the learned metric can rank the training samples with a large margin, it can rank the deviation score of unseen anomalies and background correctly. To satisfy the condition, we instantiate the deviation metric as a learned model and optimize it with proposed pixel-level and feature-level large-margin losses. The pixel-level loss is derived directly from the classical ranking metric AUC, where the discrete zero-one loss is replaced with the designed differentiable log loss. The feature-level loss optimizes the deviation ranking of extracted features in an equivalent way, which enlarges the distance of the enclosing hypersphere centers between the anomaly and background features. With simulated anomalies, both pixel-level and feature-level optimization work together to learn the transferring deviation metric, which is validated with five remote sensing modalities.

Focusing on the deviation learning target, this study instantiates the learnable deviation metric with a simple multi-scale convolutional network. Some potentially useful technologies such as transformer block, large-scale self-supervising are not used. Besides, feature-level and pixel-level ranking losses were found not to be completely mutually beneficial at the training stage (as in Fig. 14), implying the simple weighting method can be further improved.

UniADRS has unified the anomaly detection task for different modalities. Despite this, anomaly detection is the first step to extract the potential targets and the detectors cannot distinguish between real anomalies and detections that are not of interest. The latter recognition step is necessary for practical application [33]. To date, few studies have tried to combine the tasks and construct a complete detection and recognition pipeline. Leveraging the zero-shot anomaly detection ability of UniADRS and the zero-shot recognition ability of many foundation models to construct the complete "detection-recognition" pipeline is our next goal.

## ACKNOWLEDGMENT

This work was supported by the National Natural Science Foundation of China under Grant No.42325105.